\documentclass[journal]{IEEEtran}
\ifCLASSINFOpdf
  \usepackage[pdftex]{graphicx}
  \usepackage[colorlinks=true, linkcolor=blue, urlcolor=blue, citecolor=blue]{hyperref}
  \graphicspath{{../pdf/}{../jpeg/}}
  \DeclareGraphicsExtensions{.pdf,.jpeg,.png}
\else
\fi
\usepackage{url}

\begin{document}
%
\title{\huge CERBERUS: Crack Evaluation \& Recognition Benchmark for Engineering Reliability \& Urban Stability}
%
%
%



\author{Justin Reinman\textsuperscript{1} and Sunwoong Choi\textsuperscript{2}\\

\thanks{$^{1}$Palisades Charter High School, Pacific Palisades, CA, USA.}%
\thanks{$^{2}$Mechanical and Aerospace Engineering, UCLA, CA, USA.}%
}

\maketitle




%
\IEEEpeerreviewmaketitle

\begin{abstract}
CERBERUS is a synthetic benchmark designed to help train and evaluate AI models for detecting cracks and other defects in infrastructure. It includes a crack image generator and realistic 3D inspection scenarios built in Unity. The benchmark features two types of setups: a simple Fly-By wall inspection and a more complex Underpass scene with lighting and geometry challenges. We tested a popular object detection model (YOLO) using different combinations of synthetic and real crack data. Results show that combining synthetic and real data improves performance on real-world images. CERBERUS provides a flexible, repeatable way to test defect detection systems and supports future research in automated infrastructure inspection. CERBERUS is publicly available at \url{https://github.com/justinreinman/Cerberus-Defect-Generator}.
\end{abstract}

\section{Introduction}
%
%
%
%
\IEEEPARstart{I}{nfrastructure} deterioration is a critical issue, with 7.5\% of bridges in the United States classified as being in poor condition in 2021 \cite{intro:bridgestat}. Urban areas are particularly vulnerable to the consequences of failing infrastructure despite serving as vital economic hubs, contributing about 91\% of the United States' gross domestic product (GDP) and containing trillions of dollars in land value. The five largest cities alone accounted for 23\% of the nation's GDP, highlighting their economic importance. However, aging and deteriorating infrastructure poses a significant risk, with projections estimating that neglecting necessary repairs and replacements could cost the U.S. GDP up to \$10 trillion by 2039 \cite{intro:cost}. This estimate includes losses from reduced goods production, disrupted supply chains, fewer jobs, and lower wages, while also affecting essential services such as transportation, water, electricity, and sanitation -- ultimately harming public health and quality of life \cite{intro:EPA}.

Timely visual inspections are crucial for assessing structural issues such as cracks, spalling, and corrosion, which can compromise the integrity of these structures if left unaddressed \cite{intro1}\cite{intro2}. However, traditional inspection methods can be labor-intensive, costly, and difficult to conduct in remote or hazardous locations. To address these challenges, drones have been increasingly utilized to assist in structural assessments \cite{drones}.

Machine learning presents an opportunity to automate defect detection in these drone inspections, often functioning alongside human supervisors to improve efficiency and accuracy \cite{Choi24}. While prior research \cite{crack1bib}-\cite{crack4bib} has explored machine learning applications for crack detection, many studies rely on ad-hoc small datasets, making it difficult to compare techniques or evaluate their real-world performance. The lack of standardized benchmarks hinders progress in developing more robust and scalable solutions. What is needed is a repeatable yet realistic benchmark suite to provide a reliable comparison point for evaluating machine learning-based defect detection techniques.

We chose to create a synthetic suite because of the flexibility afforded over real world recording. It should be used as an intermediate step in training a machine model, before ultimately tested in a real world environment, as we demonstrate in this paper.

In particular, we propose a synthetic benchmark suite that provides:
\begin{itemize}
    \item a defect recognition framework for repeatable visual scenarios
    \item the flexibility to create different structural materials, defect placement, and challenging distractions
    \item a full suite of sample scenarios that challenge existing objection detection infrastructures
\end{itemize}

\section{Simulation Setup}
In the construction of CERBERUS, we created a crack generator, designed visual scenarios in Unity, and ultimately tested the benchmark suite using YOLO.

\subsection{Crack Generator}
Machine learning relies on large amounts of robust training data. Rather than only using a database of existing cracks, we created a pixel-by-pixel crack generator written in C to provide diverse yet realistic training data for defect detection models. By procedurally generating cracks, we can vary features such as branching, thickness, coloration, and direction, ensuring a wide range of variations for our learning models. 

\begin{figure}
    \centering
    \includegraphics[width=0.20\linewidth]{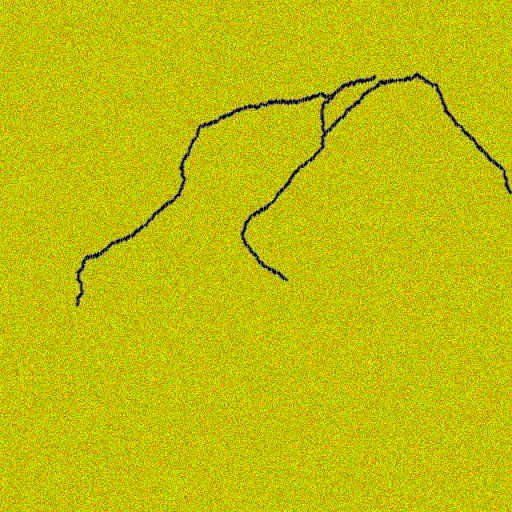}
    \includegraphics[width=0.20\linewidth]{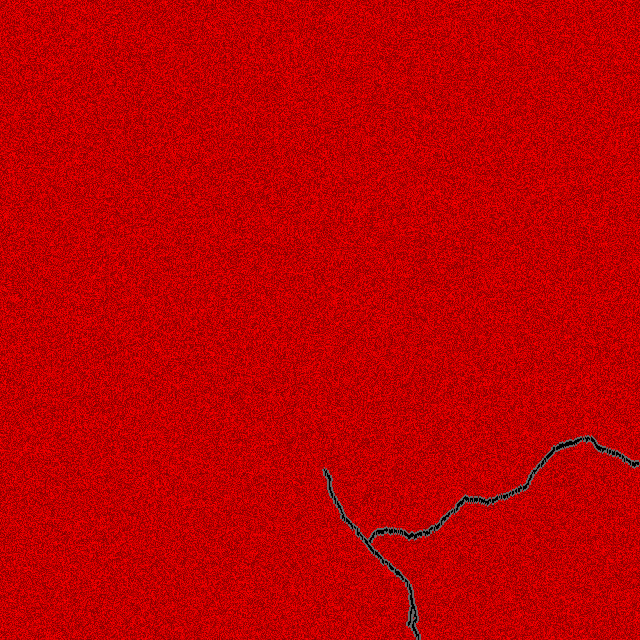}
   \includegraphics[width=0.20\linewidth]{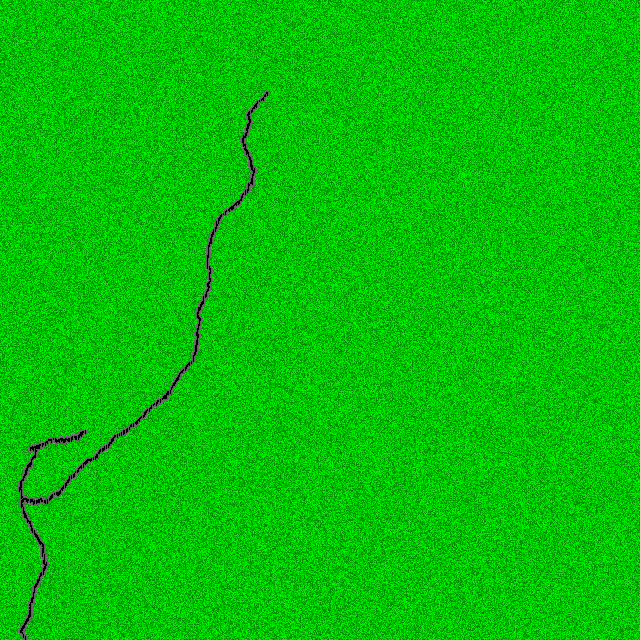}
   \includegraphics[width=0.20\linewidth]{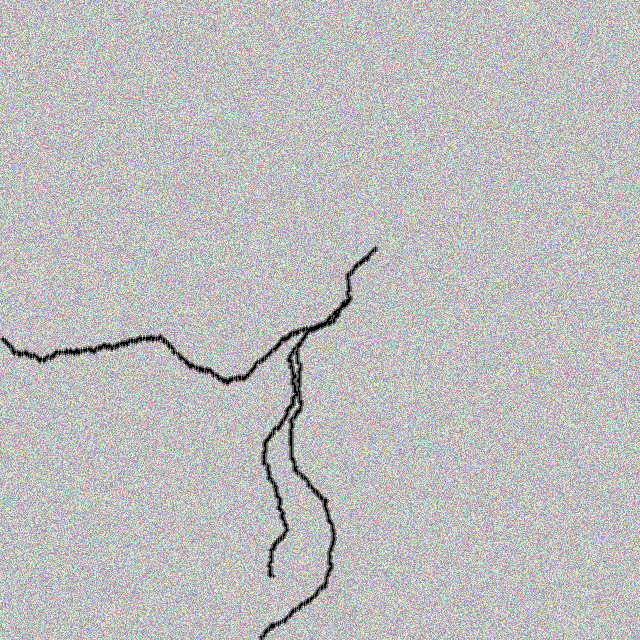}
    \caption{Exemplary cracks generated for CERBERUS}
    \label{fig:crack-examples}
\end{figure}

We built the crack generator on top of the libattopng \cite{libatt} C library for uncompressed png generation. The crack generator creates 640 x 640 pixel images with randomized color values. The crack generator picks a random starting point in the image and a random starting direction. The crack generator colors 2-4 pixels at that location (depending on the randomly selected crack thickness), and then continues to draw the hairline crack pixel by pixel in the current direction. The main loop of the crack generator randomly inserts pertubations (small movements in horizontal or vertical directions) and changes in direction to model the jagged nature of hairline cracks. The generator randomly determines the overall length of the crack and once that length has been reached, the model may randomly generate offshoots (branches) from the crack. The crack is eventually bordered by a gradient of color values to create a more realistic transition to the crack, modeling a wearing away of material at the edges of the crack. Figure \ref{fig:crack-examples} demonstrates four cracks generated as part of the training set.

\subsection{Unity}

Unity \cite{unitybib} is a development platform that provides a flexible environment for creating interactive 3D applications. In our project, we utilized Unity 2022.3 to develop CERBERUS, leveraging its rendering pipeline and shader capabilities to create realistic defect simulations. We used the dolly camera in Unity to create predefined drone travel paths through the virtual environment. Additionally, Unity enabled efficient video dumping, allowing us to capture high-quality footage of defect visualizations for further analysis and machine learning training. This functionality was essential for generating a dataset of defect examples, ensuring consistency and reproducibility in our benchmarking process. We used Unity's HD Render Pipeline and videos were created using Unity's Recorder in FHD 1080p at a 16:9 aspect ratio. 

\subsection{You Only Look Once (YOLO)}
YOLO \cite{yolobib} is an object detection algorithm optimized for real-time performance. It employs a single neural network to simultaneously predict object locations and class probabilities, enabling fast and efficient image analysis for real-time applications. We used YOLO11 for defect recognition in our evaluation, training for 200 epochs on different data sets with separate training and validation sets to assess model performance. 

Our crack generator outputs training, validation, and testing image files and YOLO formatted labels. For the synthetic data used in this study, the crack generator labels the data by automatically finding the bounding box around the randomly generated crack. For the real world data used in this study, we created labels manually. For this study, we used a split of 80\% training data, 10\% validation data, and 10\% testing data.

\section{CERBERUS}

Benchmark scenarios provided by CERBERUS can be divided into two categories based on inspection difficulty: (1) the Drone Fly-By Scenario (Section A), and (2) the Underpass Scenario (Section B). The Drone Fly-By Scenario represents simple wall visual inspection tasks, where a drone moves laterally to scan the wall surface. In contrast, the Underpass Scenario involves box culvert inspections, characterized by varying lighting conditions and a broader range of defect locations. These two scenarios enable benchmarking of AI models for crack detection under realistic visual inspection conditions.

\subsection{Drone Fly-By Scenario}
The Fly-By portion of CERBER simulates a drone flying parallel to a section of wall, performing a visual inspection. The wall is populated with randomly placed defects and distractions. The wall material (e.g. texture to visually create the wall) and the number of defects/distractions can be adjusted to create various testing scenarios. Cracks are sourced from the test directory of the crack generator output, with other defects applied as decals, representing images of common masonry issues. Considering a realistic scenario, distractions are sourced from a directory of images that include movie posters and graffiti art \cite{graffiti_generator},
and are included to evaluate the model's ability to differentiate between structural defects and irrelevant visual noise. In addition to providing the Unity model to generate custom Fly-By benchmarks, CERBERUS includes eight pre-generated Fly-By Scenarios to exercise an object recognition framework, varying the material of the wall, the number of defects on the wall, and the number of distractions on the wall (as summarized in Table \ref{tab:flyby}). 

\begin{table}[h]
    \centering
    \caption{Pre-Generated Drone Fly-By Scenarios in CERBERUS}
    \label{tab:flyby}
    \begin{tabular}{|c|c|c|}
    \hline
    Material & \# of Defects & \# of Distractions \\
    \hline
 Brown Terracota, Dark Concrete, & 2 or 3  & 0 or 2   \\
    Light Concrete, Seamed Concrete   & & \\
    \hline
    \end{tabular}
\end{table}

\subsection{Underpass Scenario}
In this more complex scenario, the drone navigates a fixed path through a structured environment with varying geometry and shadows. The scene includes a predetermined set of defects and distractions to ensure repeatable testing. Structural complexity and lighting variations challenge the model’s ability to accurately detect defects while filtering out irrelevant visual noise. This setup provides a future bar for object recognition systems to achieve. 

\section{Sample Evaluation}
As an example of use, we tested our benchmarking methodology with an object detection model that was trained to detect hairline cracks in concrete. We did not train our object detector to avoid distractions or discover other types of defects. We contrast the use of synthetic and real world data in our object detection approach in the next section.

\subsection{Methodology}
Our study trained YOLO with crack detection models using three main data configurations: (1) \textit{Synthetic Only} - generated training data only with our crack generator; (2) \textit{Real Only} - generated training data only with real world photos of cracks; and (3) \textit{Synthetic and Real} - a hybrid approach using data from both our crack generator and the real world photos of cracks. Table \ref{tab:trainingdata} summarizes the training data used in our study. \textit{Synthetic and Real} uses the same training and validation sets from both \textit{Real Only} and \textit{Synthetic Only} combined. In addition to the basic three configurations, we used additional real world training and validation data for \textit{Real Only (More Training Data)} to test the benefit of additional training data. Similarly, we used additional synthetic training and validation data for \textit{Synthetic and Real (More Training Data)}. All configurations were trained for 200 epochs with YOLO11.

\begin{table}[h]
    \centering
    \caption{YOLO Training Configurations}
    \label{tab:trainingdata}
    \begin{tabular}{|c|c|c|}
    \hline
    Training & Training & Validation  \\
    Data & Samples & Samples \\
    \hline
    \hline
    Real Only & 101 & 19\\
    \hline
    Real Only & 321 & 91 \\
    (More Training Data) &  & \\
    \hline
    Synthetic Only & 180 & 40 \\
    \hline
    Synthetic and Real & 281 & 59\\
    \hline
    Synthetic and Real & 861 & 152 \\
    (More Training Data) &  & \\
    \hline
    \end{tabular}
\end{table}

\subsection{Training Results}
\begin{figure}
    \centering
    \includegraphics[width=1\linewidth]{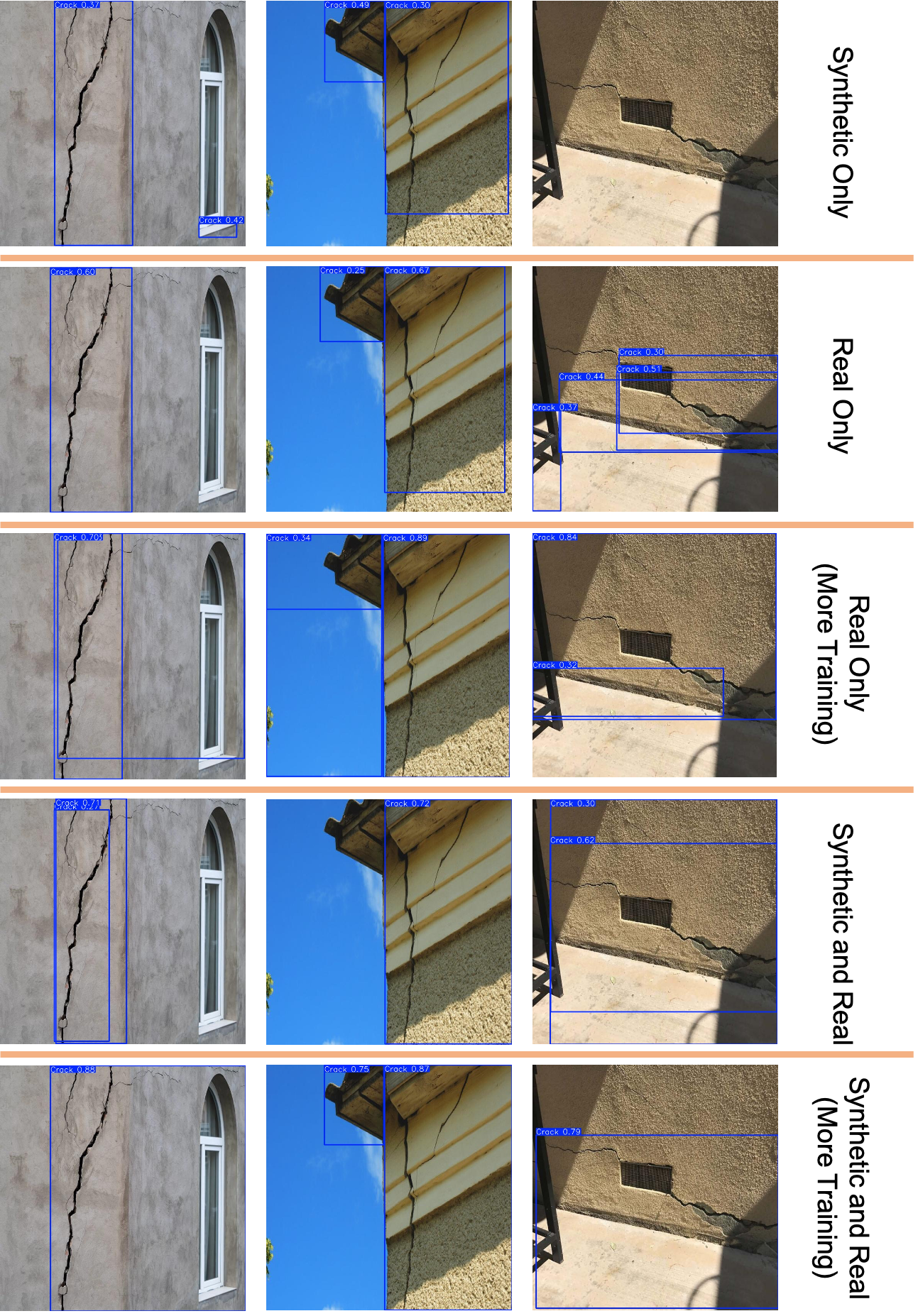}
    \caption{Object detection results on real-world images for the five configurations from Table \ref{tab:trainingdata}}
    \label{fig:compare-1}
\end{figure}

Figure \ref{fig:compare-1} shows exemplary results from three real world images showing cracks on the five different configurations. For the first column, there is the larger crack on the left portion of the image with a branching smaller crack that extends to the next perpendicular wall along the top of the window. \textit{Synthetic Only}, \textit{Real Only}, and \textit{Synthetic and Real} detect the larger crack, but not the smaller one.  \textit{Real Only (More Training Data)} is able to detect both cracks, but as separate overlapping bounding boxes. \textit{Synthetic and Real (More Training Data)} is able to detect the entire crack system as a single bounding box. 
In the second column, the \textit{Synthetic Only} and \textit{Real Only} performed similarly, missing the bottom portion of the crack with a reasonable amount of certainty. \textit{Real Only (More Training)} and \textit{Synthetic and Real (More Training)} performed with greater certainty than before, managing to capture the bottom parts of the crack, but failing to realize that the shingles were not a crack. \textit{Synthetic and Real} performed the best, capturing the entirety of the crack and noting that the shingle was not a crack.

Lastly, in the final column, \textit{Synthetic Only} failed to notice the crack at all. \textit{Real Only} noted a portion of the crack with multiple overlapping detections, but also incorrectly identified a chair leg as a crack. \textit{Real Only (More Training)} missed a larger portion of the crack, and incorrectly identified portions of the textured wall at the top of the image where there is no crack. {Synthetic and Real} identified the full crack, as well as the cracks in the concrete floor, but also identified portions of the texture wall at the top of the image as a false positive crack.  \textit{Synthetic and Real (More Training)} performed the best, correctly identifying the full crack network (both wall and floor) with higher levels of certainty.

\begin{figure}
    \centering
    \includegraphics[width=1\linewidth]{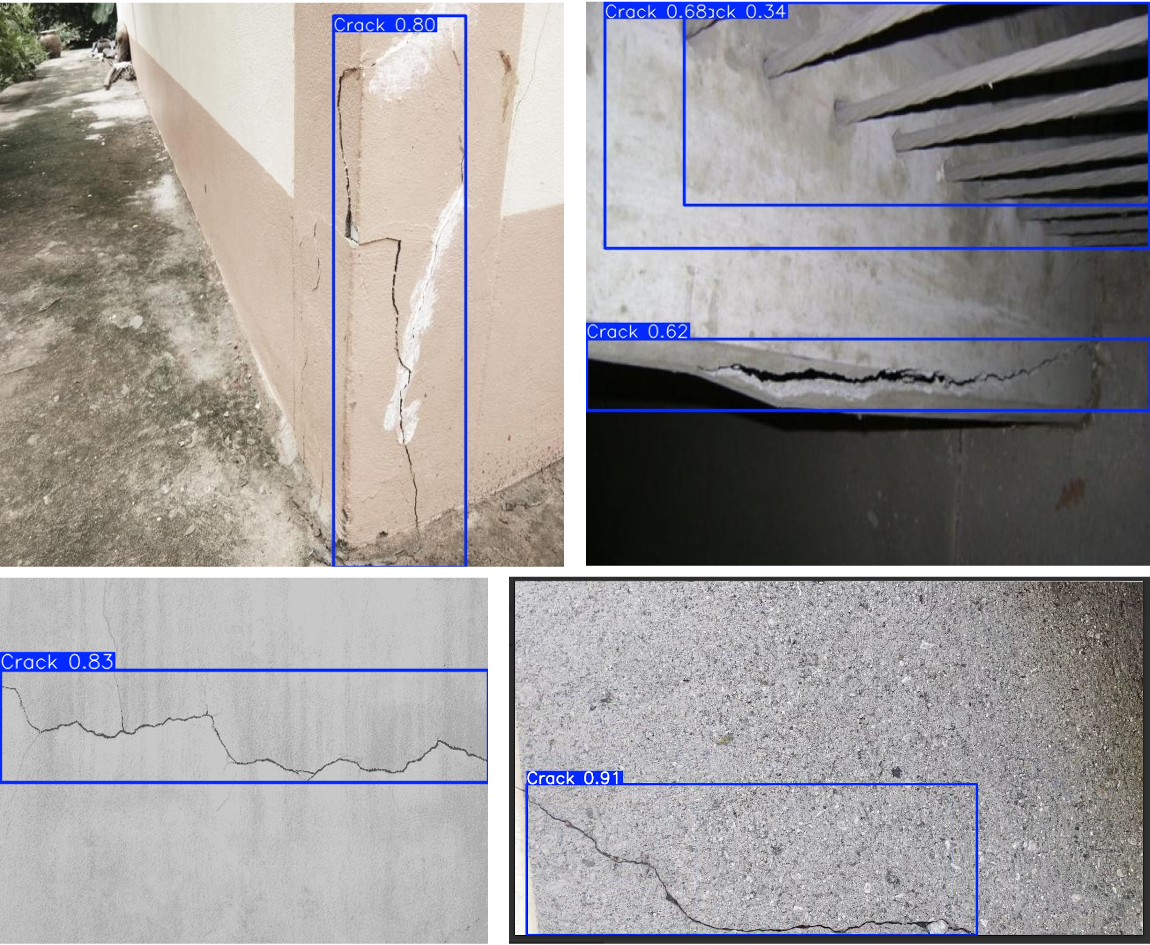}
    \caption{Additional object detection results for \textit{Synthetic and Real (More Training Data)}}
    \label{fig:compare-3}
\end{figure}

Figure \ref{fig:compare-3} shows more real world image recognition results with \textit{Synthetic and Real (More Training Data)}. In all cases, cracks are successfully detected and bounded. In the upper right portion of the figure, the piping is misidentified as cracks because the object recognition model has not been trained on piping. However, no cracks are missed.

Overall, the success of \textit{Synthetic and Real (More Training Data)} provides confidence that synthetic training data is a beneficial approach to creating robust object recognition frameworks that can work well on real world data.

\subsection{Benchmark Results}
In this section, we examine results from our \textit{Synthetic and Real (More Training Data)} model on the CERBERUS suite, as a demonstration of how this benchmarking suite could be used for future defect recognition research.

\subsubsection{Fly-By Results}
For all wall textures in the Fly-By benchmarks in CERBERUS, \textit{Synthetic and Real (More Training Data)} successfully recognized all cracks. We did not train using any of the cracks in the Benchmark suite - the cracks in CERBERUS were not the same as any of the cracks used in training or validation data for \textit{Synthetic and Real (More Training Data)}. 

\begin{figure}
    \centering
    \includegraphics[width=1\linewidth]{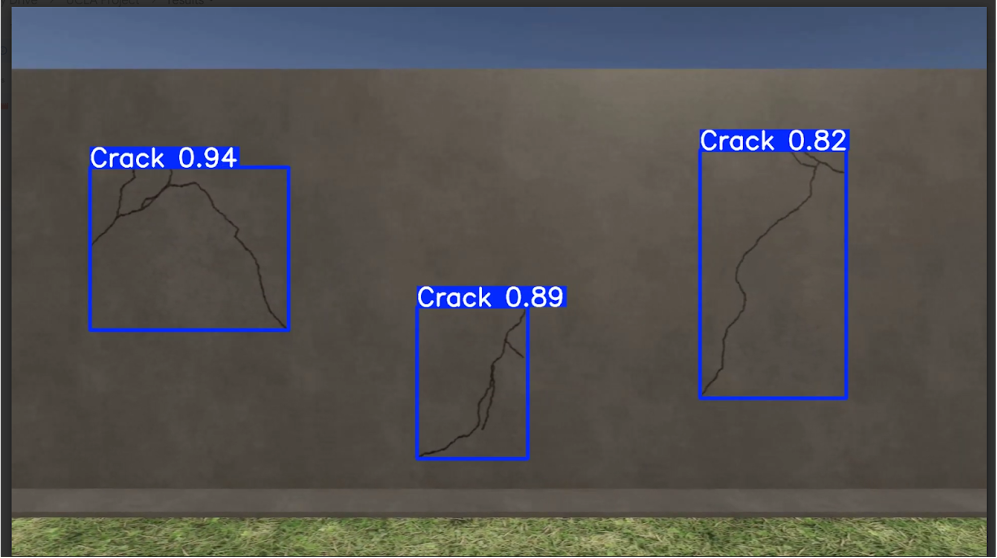}
    \caption{\textit{Synthetic and Real (More Training Data)} CERBERUS correct identification}
    \label{fig:fig_flyby_example4}
\end{figure}

For example, Figure \ref{fig:fig_flyby_example4} shows one frame from the object detection video stream of our \textit{Synthetic and Real (More Training Data)} configuration on one of our Fly-By benchmarks. 

\begin{figure}
        \centering
        \includegraphics[width=1\linewidth]{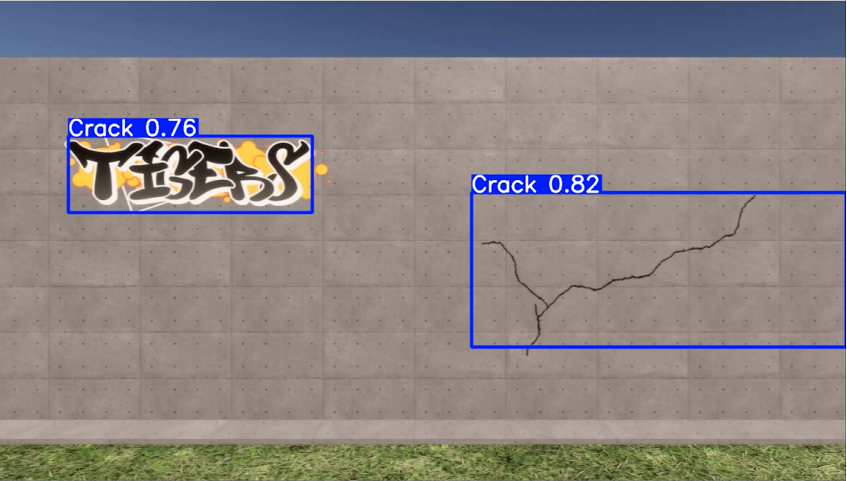}
        \caption{\textit{Synthetic and Real (More Training Data)} incorrectly identifying graffiti}
        \label{fig:fig_flyby_example2}
    \end{figure}

However, since it was not trained on distractions, elements such as posters and graffiti were sometimes misclassified. Future data sets should include more real-world noise for better robustness. Figure \ref{fig:fig_flyby_example2} shows this issue, as the program incorrectly identifies the graffiti as a crack with high certainty.\textit{Synthetic and Real (More Training Data)}

\begin{figure}
    \centering
    \includegraphics[width=1\linewidth]{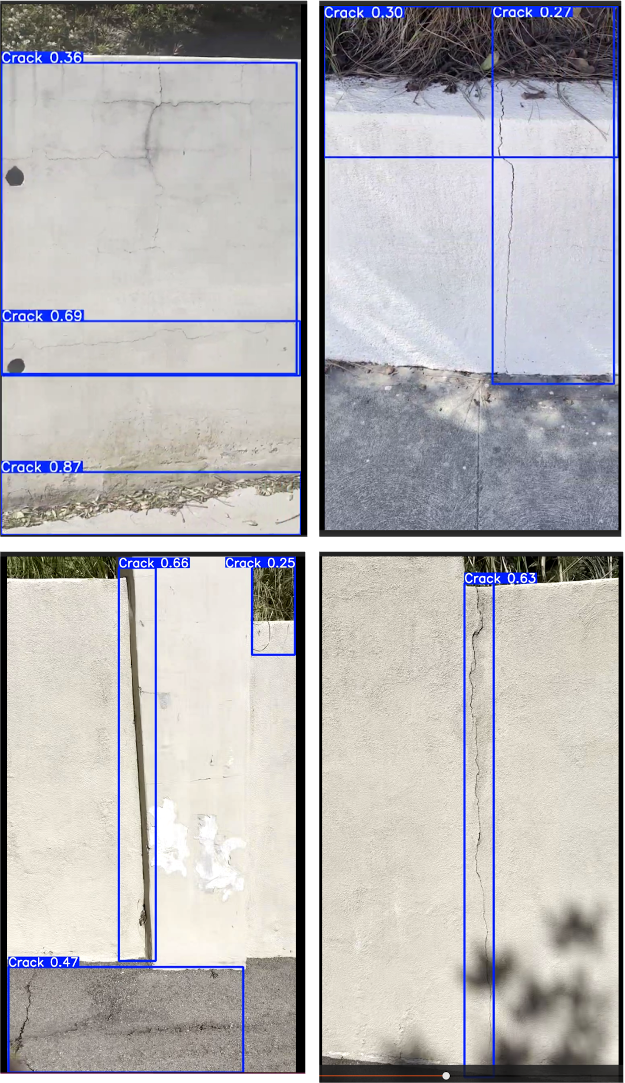}
    \caption{Real world Fly-by examples with \textit{Synthetic and Real (More Training Data)} detection}
    \label{fig:fig_flyby_realexample}
\end{figure}

As an additional comparison point, we recorded real world Fly-By video data. We walked past real world walls with hairline cracks while recording the wall on a cell phone. The resulting video was a real world version of our benchmarking suite. \textit{Synthetic and Real (More Training Data)} was able to detect the cracks in these real world videos as well, as illustrated in Figure \ref{fig:fig_flyby_realexample}. Noteably, some false positives were detected, especially at the boundaries between the wall and foliage. 

\subsubsection{Underpass Results}
\begin{figure}
    \centering
    \begin{tabular} {c}
    \includegraphics[width=1\linewidth]{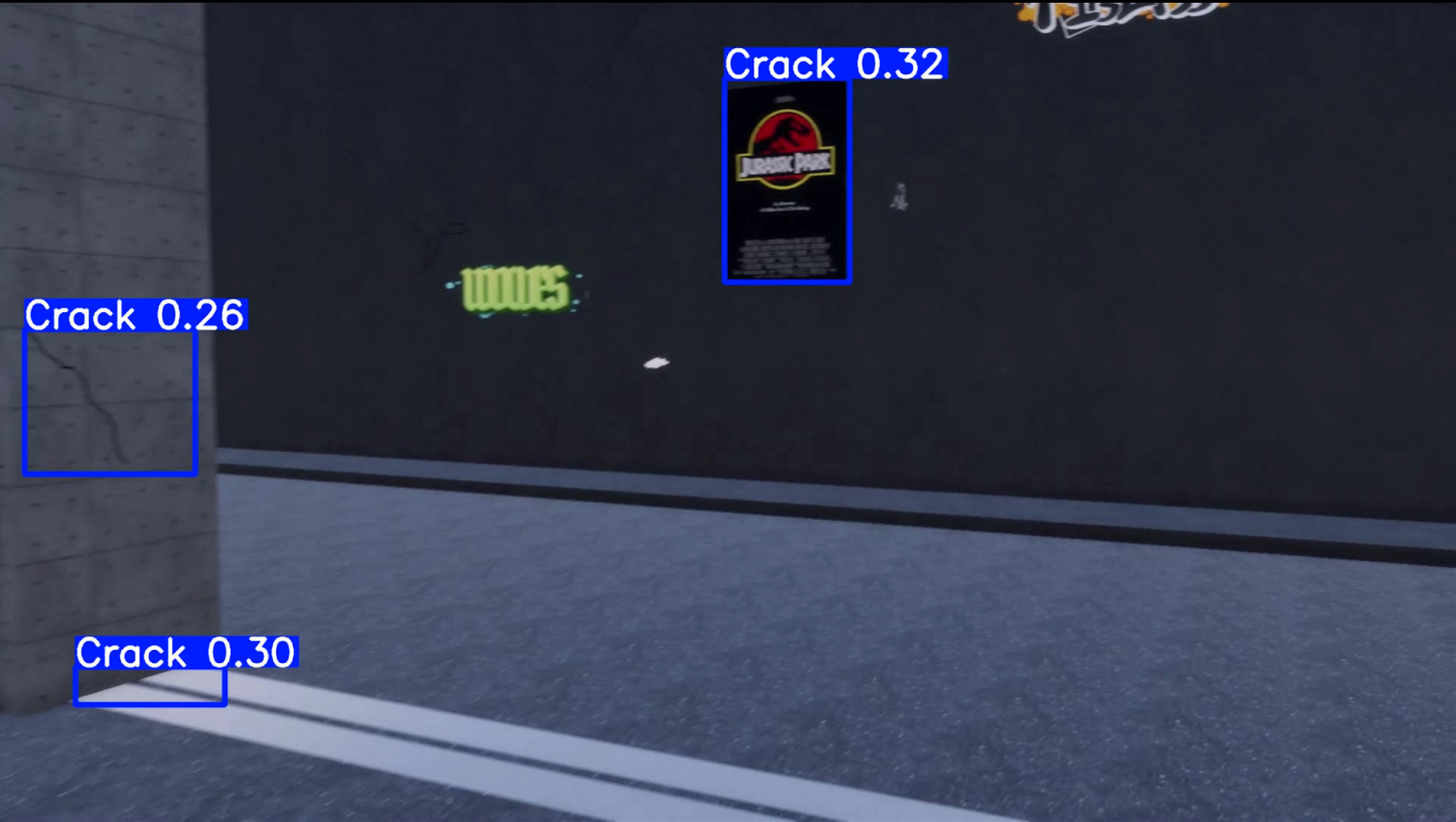}\\
    (a)\\
    \includegraphics[width=1\linewidth]{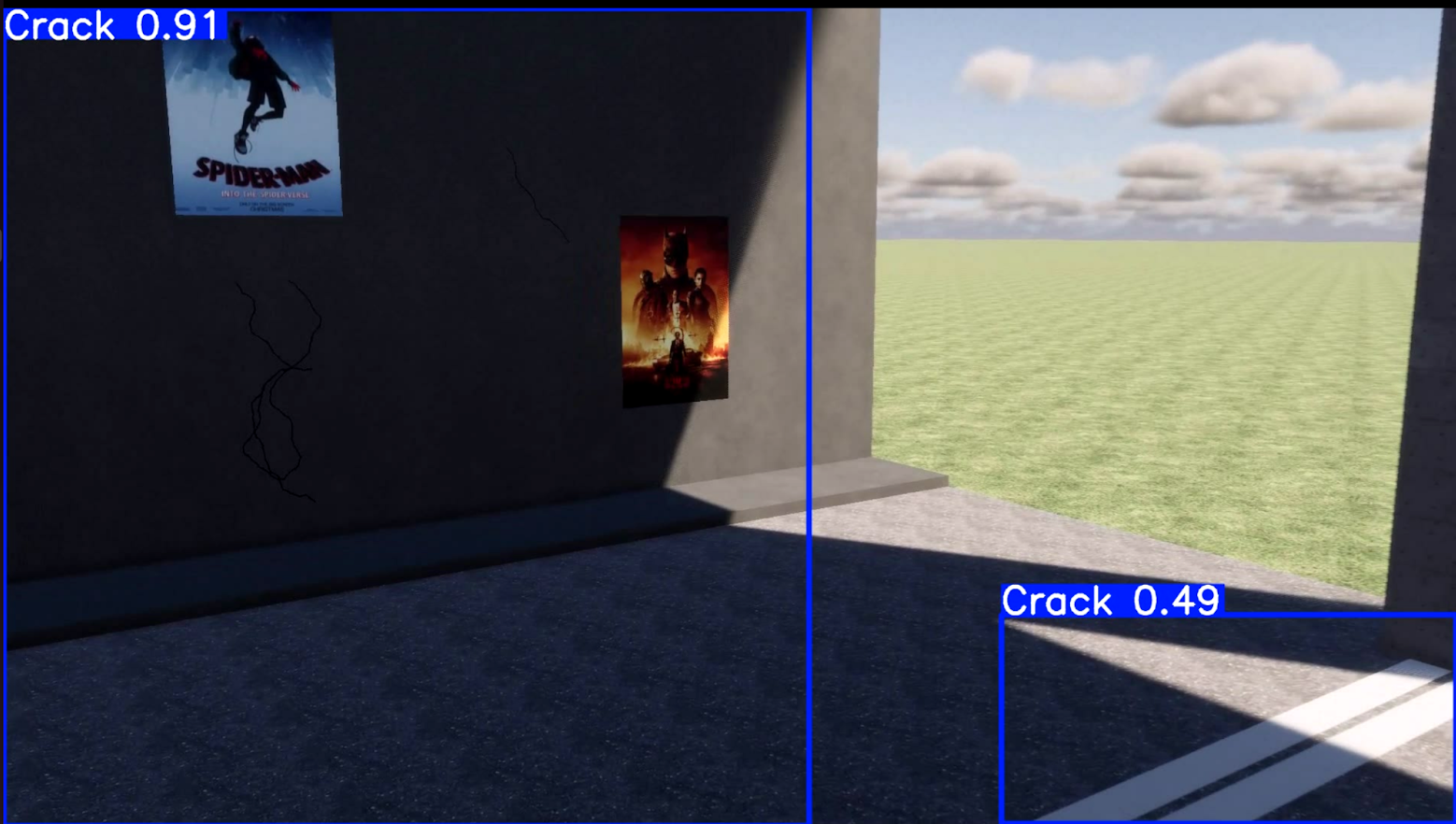}\\
    (b)\\
    \end{tabular}
    \caption{Exemplary video frames (a) and (b) from our Underpass benchmark showing the results of object detection using the \textit{Synthetic and Real (More Training Data)} configuration}
    \label{fig_underpass_example0}
\end{figure}

The Underpass Benchmark in CERBERUS was much more difficult for our \textit{Synthetic and Real (More Training Data)} configuration. Figure \ref{fig_underpass_example0} shows exemplary video frames from our results. With varied lighting, complex geometry, varied angles with different surfaces (as opposed to a more linear trajectory on the Fly-By benchmarks), and more distractions, the \textit{Synthetic and Real (More Training Data)} model detected many of the cracks but not all and often found false positives in legitimate seams (e.g. Figure \ref{fig_underpass_example0}(b)), shadows, and other distractions. 

\section{Conclusion}

CERBERUS is a forward thinking benchmarking suite for use by researchers in object recognition for structural defects. We use a synthetic crack generation framework to create a wide array of defects for the suite. Our exemplary results suggest that the use of both real world and synthetic training data can provide strong generalization for robust crack detection. Future work should focus on more complex scenarios, like our Underpass Benchmark, to be able to handle recognition in low-light and high-clutter settings.

\section*{Availability}
The crack generator and sample videos for CERBERUS are available at \url{https://github.com/justinreinman/Cerberus-Defect-Generator}.

\section*{Acknowledgment}
The authors acknowledge the support provided by Professor Narasimhan and the Sensing and Robotics Lab at UCLA.

\ifCLASSOPTIONcaptionsoff
  \newpage
\fi

\end{document}